\documentclass{article}

\usepackage{PRIMEarxiv}

\usepackage[utf8]{inputenc} 
\usepackage[T1]{fontenc}    
\usepackage{hyperref}       
\usepackage{url}            
\usepackage{booktabs}       
\usepackage{amsfonts}       
\usepackage{nicefrac}       
\usepackage{microtype}      
\usepackage{lipsum}
\usepackage{xcolor}
\usepackage{soul}
\usepackage{amsmath}
\usepackage{fancyhdr}       
\usepackage{graphicx}       
\usepackage{ifluatex}
\usepackage{subfig}
\usepackage{comment}
\usepackage{hyperref}

\ifluatex
  \usepackage{pdftexcmds}
  \makeatletter
  \let\pdfstrcmp\pdf@strcmp
  \let\pdffilemoddate\pdf@filemoddate
  \makeatother
\fi

\usepackage[notransparent]{svg}

\graphicspath{{media/}}     

\title{Combining Public Human Activity Recognition Datasets to Mitigate Labeled Data Scarcity}

\author{Riccardo Presotto \\ \small{EveryWare Lab, University of Milan} \\ \small{Milan, Italy}  \\ riccardo.presotto@unimi.it
\And
\textbf{Sannara Ek} \\ \small{Univ. Grenoble Alpes, CNRS} \\ \small{LIG F-38000, Grenoble, France}  \\ sannara.ek@univ-grenoble-alpes.fr
\And
\textbf{Gabriele Civitarese} \\ \small{EveryWare Lab, University of Milan} \\ \small{Milan, Italy}  \\ gabriele.civitarese@unimi.it
\And
\textbf{François Portet} \\ \small{Univ. Grenoble Alpes, CNRS} \\ \small{LIG F-38000, Grenoble, France}  \\ francois.portet@imag.fr
\And
\textbf{Philippe Lalanda} \\ \small{Univ. Grenoble Alpes, CNRS} \\ \small{LIG F-38000, Grenoble, France}  \\ philippe.lalanda@imag.fr
\And
\textbf{Claudio Bettini} \\ \small{EveryWare Lab, University of Milan} \\ \small{Milan, Italy}  \\claudio.bettini@unimi.it
}

\begin{document}
\maketitle

\begin{abstract}

The use of supervised learning for Human Activity Recognition (HAR) on mobile devices leads to strong classification performances. Such an approach, however, requires large amounts of labeled data, both for the initial training of the models and for their customization on specific clients (whose data often differ greatly from the training data). This is actually impractical to obtain due to the costs, intrusiveness, and time-consuming nature of data annotation. Moreover, even with the help of a significant amount of labeled data, model deployment on heterogeneous clients faces difficulties in generalizing well on unseen data. Other domains, like Computer Vision or Natural Language Processing, have proposed the notion of pre-trained models, leveraging large corpora, to reduce the need for annotated data and better manage heterogeneity. This promising approach has not been implemented in the HAR domain so far because of the lack of public datasets of sufficient size. In this paper, we propose a novel strategy to combine publicly available datasets with the goal of learning a generalized HAR model that can be fine-tuned using a limited amount of labeled data on an unseen target domain. Our experimental evaluation, which includes experimenting with different state-of-the-art neural network architectures, shows that combining public datasets can significantly reduce the number of labeled samples required to achieve satisfactory performance on an unseen target domain.

\end{abstract}

\section{Introduction}
\label{intro}

Despite years of research and progress~\cite{gu2021survey}, the field of sensor-based Human Activity Recognition (HAR) on mobile or wearable devices remains a hot research topic. This is due to several reasons. First, the field has major social importance in the sense that it is necessary for the development of value-added services in the fields of health, well-being, or task monitoring in industry. Secondly, although constant progress can be reported, the problem is far from being solved and today solutions are difficult to be deployed in real-world scenarios~\cite{wang2019deep}. Most recent approaches are based on machine learning techniques. More precisely, supervised learning leads to strong classification performances for HAR. This, however, requires large amounts of labeled data, both for the initial training of the models and for their customization on specific clients (whose data often differ greatly from the training data). This is actually a problem because of the costs, intrusiveness, and time-consuming nature of data collection and annotation. Even with the help of a number of labeled data, models deployed on heterogeneous clients face difficulties to generalize well on unseen data. 

In other domains, such as Computer Vision or Natural Language Processing, learning a model on one large dataset~\cite{russakovsky2015imagenet,evain:hal-03407172} makes it possible to transfer knowledge to new domains using only a limited amount of specific data~\cite{liu2021self}.
However, in the HAR domain, transferring knowledge from a single dataset is not effective because of the lack of publicly available large and heterogeneous datasets, due to the above-mentioned problems. We call this the labeled data scarcity problem. Ideally, a pre-trained model should be able to cope with the variability of devices~\cite{10.1145/2809695.2809718}, the variability of applications, and also the behavioral differences of users~\cite{weiss2012impact}. In fact, there are a number of public datasets but they are generally small and acquired in scripted scenarios (e.g. Mobiact~\cite{vavoulas2016mobiact}, MotionSense~\cite{Malekzadeh:2018:PSD:3195258.3195260}, PAMAP2~\cite{reiss2012introducing}). The few ones that are relatively large and acquired in the wild often include noisy data and incorrect annotations (e.g. ExtraSensory~\cite{vaizman2017recognizing}).



\begin{figure}[h!]
\centering        \includegraphics[width=8cm]{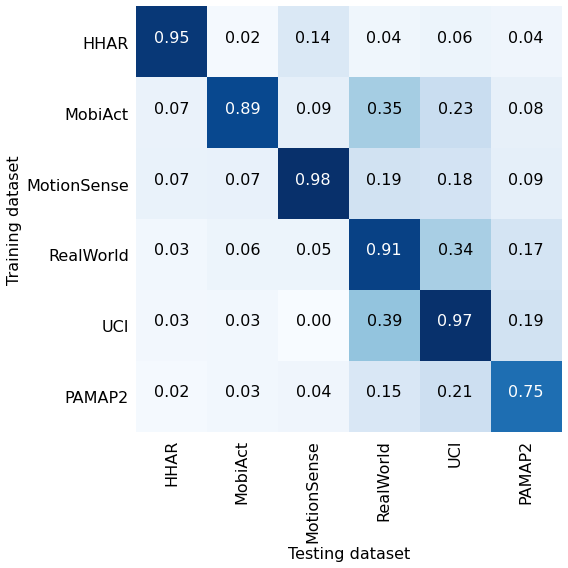}    \caption{Cross-dataset evaluation by using a state-of-the-art classifier~\cite{ronald2021isplinception}. A cell at coordinates $i,j$ shows the F1 score obtained by training the model with $70\%$ of dataset $i$ on $30\%$ of dataset $j$.}
\label{fig:cross_dataset} \end{figure}

Transfer learning from one dataset to another has been already attempted but existing works showed that such approaches lead to low recognition rates~\cite{qin2019cross}. To further demonstrate the problem, Figure~\ref{fig:cross_dataset} presents the result of an experiment showing the challenge of training a model on one dataset and expecting it to perform well on a different dataset. This cross-dataset evaluation was performed using some of the most commonly used publicly available datasets for HAR and using a state-of-the-art model~\cite{ronald2021isplinception}. 
The figure clearly shows that the model performs well when trained and tested on the same dataset, but its performance drastically deteriorates in cross-dataset scenarios. These results highlight the challenges associated with developing models that can generalize well across diverse datasets and underscore the need for developing more robust approaches that can effectively leverage knowledge from pre-training on large, diverse, and representative datasets. 

In this paper, we hypothesize that, by combining several public HAR datasets that are often used as benchmarks by the HAR community, it is possible to pre-train models that can be fine-tuned on target clients with a limited amount of labeled data.
Specifically, we propose a novel approach to combining datasets and we report our experience considering $6$ public datasets to generate a pre-trained model. Also, we propose an evaluation strategy called Leave-One-Dataset-Out that effectively assesses the generalization capabilities of our approach.

Our experiments show that our pre-trained model can be effectively transferred to an unseen target HAR domain using a limited amount of labeled data for fine-tuning. To the best of our knowledge, this is the first work proposing to combine several heterogeneous HAR datasets to mitigate the labeled data scarcity problem. Precisely, the contributions of this paper are the following:



\begin{itemize}      \item We pre-trained models using a combination of diverse publicly available HAR datasets. We show that such models can be effectively fine-tuned efficiently with limited data of unseen target domains. 
\item We then propose a novel realistic and challenging evaluation methodology termed Leave-One-Dataset-Out (LODO), and we use it to extensively assess the effectiveness of our approach on different state-of-the-art neural network architectures.
\item The pre-trained model, the corresponding training data, and the code are publicly available to make this research reproducible\footnote{https://github.com/getalp/SmartComp2023-HAR-Supervised-Pretraining}.
\end{itemize}




\section{Related Work}
\label{related}

Human activity recognition (HAR) using machine learning has traditionally relied on supervised learning approaches, which require a significant amount of labeled data for training to achieve high accuracy \cite{jobanputra2019human}. However, obtaining labeled activity data is often time-consuming, intrusive, costly, and hence unfeasible on a large scale \cite{CookFK13}. To overcome the scarcity of labeled samples in the target domain, domain adaptation (DA) techniques have emerged to leverage labeled samples that are available from a different source domain~\cite{farahani2021brief}. Among the many DA techniques, transfer learning has shown valuable results in various domains, including natural language processing \cite{zhuang2020comprehensive,radford2022robust}. However, adapting transfer learning to HAR is challenging due to device variability, body displacement, and behavioral differences among users, making it difficult to generate a generic model that can quickly adapt to diverse datasets \cite{cook2013transfer,wang2018deep}. Several researchers have investigated how to apply transfer learning methods to solve cross-domain HAR problems, including cross-device \cite{morales2016deep,kurz2011real}, cross-sensor-installation-position \cite{wang2018deep,chen2019cross}, and cross-persons \cite{lu2013hybrid, zhuang2020comprehensive}.
However, publicly available datasets for HAR have limited labeled samples and classes, which restricts the effectiveness of cross-domain adaptation works in this area \cite{gjoreski2019cross}. 
For instance, effective deep CNN frameworks have been proposed to enable the generation of a model that can efficiently adapt across diverse datasets \cite{khan2018scaling}, but they require that the classes between the source and the target domain overlap, hence limiting their effectiveness in real-world scenarios.
Moreover, most of the studies in this field have been conducted by transferring knowledge from one single dataset to another, resulting in performance variations depending on the compatibility among the data across the considered dataset pairs \cite{gjoreski2019cross}. Furthermore, due to the data scarcity problem in HAR, using only the data from one dataset to generate the model that needs to be transferred to another domain may not be sufficient. A possible solution to mitigate this problem is to pre-train the model on a different source domain before fine-tuning on the data-scarce target domain \cite{sanabria2020unsupervised,khan2018scaling,gjoreski2019cross,zhou2020xhar,qin2019cross}. The obtained results are promising even in the case very few data are used for fine-tuning the model in the target domain. However, this approach suffers from generalization across domains that differ from the target one. We argue that utilizing only a single dataset for the pre-training HAR models is not sufficient to find common features in the target domain, especially with the highly heterogeneous nature of smartphone data.
Another approach to expand the dataset available for training may consist of using data augmentation techniques like GANs \cite{wang2018sensorygans,chan2020unified}. GANs have been also used to facilitate cross-subject transfer learning \cite{soleimani2021cross}. However, GANs still require a significant amount of labeled samples to be trained.

In recent years, self-supervised learning (SSL) approaches have gained attention as a general framework for learning from unlabeled data through a pretext task \cite{jaiswal2020survey,liu2021self}. SSL has also been applied to sensor-based HAR, leading to promising results \cite{tang2021selfhar,haresamudram2022assessing}. However, even a good unsupervised or SSL model that permits adaptation using a short amount of data still requires a significant amount of data, as exemplified in other domains such as vision and speech \cite{baevski2020wav2vec}. In the HAR domain, even non-annotated data is difficult to obtain.

\section{Combining datasets for model pre-training}
\label{sec:combination}

\begin{table*}[h]
\footnotesize
\caption{Summary of datasets characteristics}
\label{tab:datasets}
\centering
\resizebox{\columnwidth}{!}{
\begin{tabular}{@{}c|c|c|c|c|c|c@{}}
\toprule
\textbf{Dataset} &
  \textbf{\begin{tabular}[c]{@{}c@{}}\# of \\ samples\end{tabular}} &
  \textbf{\begin{tabular}[c]{@{}c@{}}\# of \\ users\end{tabular}} &
  \textbf{Adopted Devices} &
  \textbf{Sampling rate} &
  \textbf{Device position} &
  \textbf{Activities} \\ \midrule
HHAR &
 85,567 &
  9 &
  \begin{tabular}[c]{@{}c@{}}Smartphones: Samsung Galaxy S3 mini,\\  Samsung Galaxy S3, LG Nexus 4,\\  Samsung Galaxy S+\\ \\ Smartwatches: LG watches, \\ Samsung Galaxy Gears\end{tabular} &
  \begin{tabular}[c]{@{}c@{}}from 50 Hz \\ to 200 Hz\end{tabular} &
  \begin{tabular}[c]{@{}c@{}}Smartphones: Waist \\ \\ Smartwatches: Wrist\end{tabular}
  &
  \begin{tabular}[c]{@{}c@{}}Biking, Sitting, Standing,\\ Walking, Upstairs, Downstairs\end{tabular} \\ \midrule
MobiAct &
  18,634 &
  61 &
  Samsung Galaxy S3 &
  \begin{tabular}[c]{@{}c@{}}from 50 Hz\\  to 200 Hz\end{tabular} &
  \begin{tabular}[c]{@{}c@{}}Waist\end{tabular} &
  \begin{tabular}[c]{@{}c@{}}Standing, Walking, Jogging, \\ Jumping, Upstairs, Downstairs, \\ Sitting, Car step in, Car step out\end{tabular} \\ \midrule
MotionSense &
  17,231  &
  24 &
  Apple iPhone 6s &
  50 Hz &
  \begin{tabular}[c]{@{}c@{}}Waist\end{tabular} &
  \begin{tabular}[c]{@{}c@{}}Downstairs, Upstairs, Sitting, \\ Standing, Walking, Running\end{tabular} \\ \midrule
RealWorld &
  356,427 &
  15 &
  \begin{tabular}[c]{@{}c@{}}Samsung Galaxy S4 \\ \\ LG G Watch R\end{tabular} &
  50 Hz &
  \begin{tabular}[c]{@{}c@{}}Smartphones: Head, Chest, Upper arm, \\ Waist, Thigh, Shin  \\ \\ Smartwatches: Forearm\end{tabular} &
  \begin{tabular}[c]{@{}c@{}}Downstairs, Upstairs, Lying, \\ Sitting, Standing, Jumping, \\ Walking, Running\end{tabular} \\ \midrule
UCI &
  10,299 &
  30 &
  Samsung Galaxy S II &
  50 Hz &
  Waist &
  \begin{tabular}[c]{@{}c@{}}Walking, Upstairs, Downstairs, \\ Sitting, Standing, Lying\end{tabular} \\ \midrule
PAMAP2 &
  15,177 &
  8 &
  Colibri wireless IMU sensors &
  100 Hz &
  Waist, Chest, Wrist &
  \begin{tabular}[c]{@{}c@{}}Rope Jumping, Lying, Sitting,\\  Standing, Walking, Running, \\ Cycling, Nordic walking, \\ Upstairs, Downstairs,\\ Vacuum cleaning, Ironing\end{tabular} \\ \bottomrule
\end{tabular}
}

\end{table*}

In this section, we outline our strategy for combining datasets to create a pre-trained model that can generalize well across all datasets. We begin by providing an overview of the datasets we considered for this task. Next, we describe our proposed approach for combining these datasets to generate the pre-trained model.

\subsection{Datasets}

In the following, we describe the datasets we considered in this work. We note here that we explicitly seek datasets with both Accelerometer and Gyroscope data, as utilizing the two sensors together gave the best performance \cite{ek2022lightweight}. Information on the datasets is summarized in Table~\ref{tab:datasets}.

\subsubsection{Heterogeneity Human Activity Recognition (HHAR) \cite{10.1145/2809695.2809718} }
The HHAR dataset consists of $4.5$ hours of recorded activities from $9$ participants. Each participant wore 8 Android smartphones in a tight pouch carried around on the waist, and 4 Android smartwatches while performing 6 different activities (Biking, Sitting, Standing, Walking, Upstairs, and Downstairs). All 12 devices recorded the activities using accelerometer and gyroscope measurements at their maximum sampling rates, which ranged from 50 Hz to 200 Hz. The HHAR dataset represents a heterogeneous learning environment due to the variety of devices used in the data collection process.

\subsubsection{MobiAct \cite{vavoulas2016mobiact} }
    \label{subsec:mobiact} This dataset includes labeled data from $61$ different subjects with high variance in age and physical characteristics. The dataset contains data from a triaxial accelerometer, gyroscope, and magnetometer embedded into a Samsung Galaxy S3 smartphone carried by users while performing $9$ physical activities.
    During the acquisition process, the users were left free to position the smartphone with a random orientation into one of their trousers' pockets.
    The physical activities included in this dataset are the following: Standing, Walking, Jogging, Jumping, Upstairs, Downstairs, Sitting, Car step in, Car step out.
    The adopted data acquisition frequency is the highest enabled by the sensors of the selected smartphone (i.e., at most $200Hz$).
\subsubsection{MotionSense \cite{Malekzadeh:2018:PSD:3195258.3195260}} 
This dataset includes data from $24$ subjects very heterogeneous in terms of gender, age, weight, and height. The six activities performed were Downstairs, Upstairs, Sitting, Standing, Walking, and Jogging/Running. The tests were conducted using an Apple iPhone 6s that was kept in the subject's front pocket and recorded data from accelerometer, attitude, and gyroscope sensors at a 50 Hz sampling rate. 

\subsubsection{RealWorld \cite{realword}}
This dataset consists of 18 hours of recorded accelerometer and gyroscope data collected in 2016 from 15 subjects using a Samsung Galaxy S4 smartphone and an LG G Watch R placed at 7 different body positions: head, chest, upper arm, waist, forearm, thigh, and shin. The sampling rate was 50 Hz. The subjects performed various activities outdoors without any restrictions, and the data was labeled into 8 activities: Downstairs, Upstairs, Lying, Sitting, Standing, Walking, Jumping, and Running. This dataset aims to simulate the class imbalance that resembles the ones of realistic datasets. For instance, the "standing" activity represents 14\% of the data while the "jumping" activity only accounts for 2\%. 

\subsubsection{UCI Human Activity Recognition \cite{Anguita2013APD} }
This dataset was collected using a Samsung Galaxy S II placed on the participant's waist, with a sampling rate of 50 Hz. The dataset comprises 3.6 hours of recorded activities from 30 participants with an age range of 19 to 48 years old. The six activities recorded are Walking, Upstairs, Downstairs, Sitting, Standing, and Lying, and the experiments were conducted in a controlled indoor lab environment.

\subsubsection{PAMAP2 \cite{reiss2012introducing}}
    \label{subsec:PAMAP}
     This dataset was recorded in a controlled setting where 9 participants carried out 12 activities of daily living, including domestic tasks and physical exercises. The activity data were recorded using three Colibri wireless IMUs (inertial measurement units) consisting of triaxial accelerometers, gyroscopes, and magnetometers attached to the participants' ankle, chest, and wrist. The activities monitored were: rope jumping, lying, sitting, standing, walking,  running, cycling, nordic walking, ascending stairs,  descending stairs,  vacuum cleaning, and ironing. The sampling rate adopted for data acquisition was set to $100 Hz$.

\subsection{Datasets combination strategy}
\label{subsec:combination}

In the following, we describe how we pre-processed and put together the above-mentioned datasets. First, each dataset was downsampled to 50 Hz to align with the optimal frequency for HAR on smartphones as suggested by a recent survey~\cite{overviewHar}. Indeed, this survey suggests that a sampling rate between 20 Hz and 50 Hz is ideal for this task, and accelerometers and gyroscopes are the most suitable sensors. Using higher frequencies beyond 50 Hz would result in increased computation and memory costs with only marginal improvement in performance.

After downsampling, each dataset was standardized individually using sensor-wise z-normalization to center the data. We avoid standardizing all datasets together as a whole to avoid datasets with lower samples being under-presented when combined with many of the larger datasets (e.g. the large HHAR vs the small UCI dataset). In order to focus on the data scarcity problem and to limit influences on the results caused by domain shifts from different positions~\cite{ek2022lightweight},
from each dataset we use data only from the `waist' position, which is usually the most common one and it is included in all of the datasets studied in this work.

Finally, based on past studies~\cite{ignatov2018real}, the data was then segmented into instances using a window length of $128$ samples (2.56 seconds) with a 50\% overlap over all 6 channels from the accelerometer and gyroscope readings.
In order to combine the datasets together, we performed a union of the physical activities involved.
This combination led to a dataset with $147$ subjects, $10$ unique activities (`Downstairs', `Upstairs', `Running', `Sitting', `Standing', `Walking',` Lying',` Cycling', `Nordic Walking', `Jumping'), and  213,289 data samples ($\approx 151$ hours of usable data), and $12$ different devices for data acquisition.
Despite considering only one on-body position, we believe that the wide variety of subjects and devices presents a challenging learning problem due to the heterogeneity of the combined data.

\subsection{Pre-training}
The set of labeled datasets can then be used to train a deep neural network model in a fully supervised learning fashion. 
Specifically, we consider the neural network as composed of two parts: the \textit{feature extractor} and the \textit{classification head}. During the pre-training phase, we train the whole network (i.e. \textit{feature extractor} and \textit{classification head}). 
We will refer to the \textit{feature extractor} trained on several joint datasets as the \textit{pre-training model}.
By doing so, we aim to leverage the diversity and heterogeneity of the joint datasets to extract high-level and general-purpose features. These features should be applicable to a wide range of unseen domains considering various users, devices, and sensors. 
In other words, the pre-trained model is designed to capture the common underlying patterns and characteristics of the human activities that are invariant across datasets, so that it provides a robust feature extractor. 

\begin{figure*}[ht]
  \centering
      \includegraphics[scale=0.15]{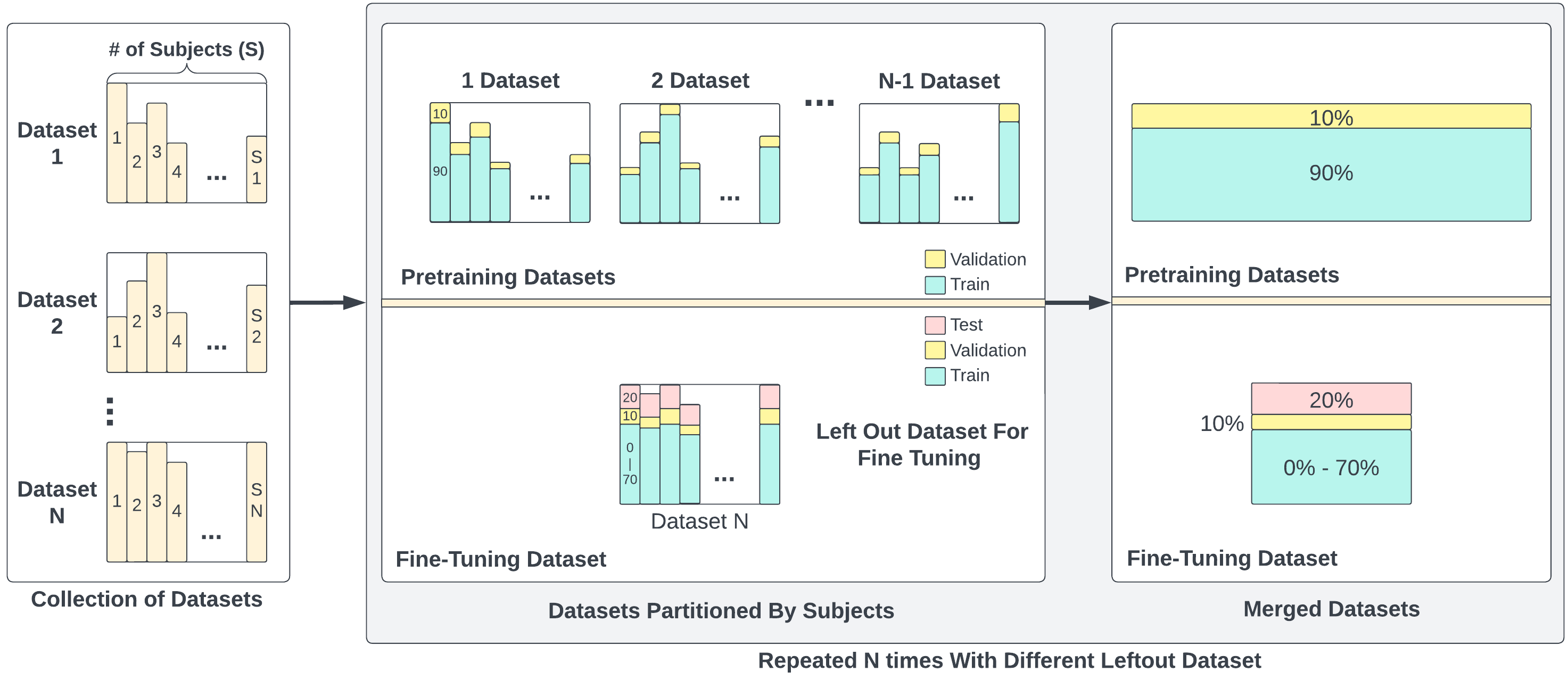}
  \caption{Overview of LODO}
  \label{fig:LODO}
\end{figure*}

\subsection{Considered Models}
In this paper, we consider four alternative state-of-the-art neural networks which we use with their default parameters and implementation from their official source code repositories. For each model, we consider the last layer (softmax classification layer) as \textit{classification head}, while the remaining layers as \textit{feature extractor}.

\begin{enumerate}
\item {\textbf{ISPL Inception} \cite{ronald2021isplinception}} 
This network has been designed by adapting the Inception-ResNet architecture proposed ~\cite{szegedy2017inception} to the HAR domain. This architecture represents a strong CNN architecture baseline for our study.

\item {\textbf{DeepConvLSTM} \cite{ordonez2016deep}} is a combination of convolutional and LSTM recurrent layers. The network architecture has been frequently used as a baseline for state-of-the-art studies in the wearable HAR domain community.

\item {\textbf{HART} \cite{ek2022lightweight}} A sensor-wise HAR Transformer (HART) architecture adapted from the successful transformer model from the vision domain \cite{NIPS2017_3f5ee243}. The model incorporates lightweight components, able to be deployed on small devices capable of real-time performance.

\item {\textbf{MobileHART}\cite{ek2022lightweight}} An extension of HART that incorporates convolutional layers to learn spatial/temporal inductive biases that conventional transformers do not. The architecture reports state-of-the-art results on multiple benchmark datasets.
\end{enumerate}

\section{Fine-tuning and evaluation}

In this section, we outline our methodology for evaluating the pre-trained model that was learned from the data obtained by combining the considered datasets.

\subsection{Fine-tuning Methodologies}
A small amount of labeled data acquired on the target domain is used to fine-tune the pre-trained model. 
For the sake of this work, in this phase, we consider three strategies for using this small amount of data with the different models composed of the pre-trained \textit{feature extractor} augmented with a randomly initialized \textit{classification head}: 
a) using an unfrozen and randomly initialized \textit{feature extractor} (denoted as $Rd$) as a baseline, b) using a frozen pre-trained \textit{feature extractor} ($P_{F}$) to assess generalization capabilities on new tasks and c) using a pre-trained feature extractor with all layers unfrozen ($P_{U}$) to assess the impact of fine-tuning data on feature extraction.
\begin{table}[ht]
\footnotesize
\centering
\caption{Fine-tuning ratios and the corresponding samples of training data for the considered datasets}
\begin{tabular}{@{}ccccc@{}}
\toprule
& \multicolumn{4}{c}{Ratios of training data used for fine-tuning } \\ \midrule
\multicolumn{1}{c|}{\textbf{Dataset}} &
  \multicolumn{1}{c|}{\textbf{   1\% }} &
  \multicolumn{1}{c|}{\textbf{  5\%  }} &
  \multicolumn{1}{c|}{\textbf{ 10\% }} &
  \multicolumn{1}{c}{\textbf{ 70\% (All Train Data) }} \\ \midrule
\multicolumn{1}{c|}{HHAR}        
& \multicolumn{1}{c|}{733} & \multicolumn{1}{c|}{3,680} & \multicolumn{1}{c|}{7,366} & \multicolumn{1}{c}{51,581} \\ \midrule
\multicolumn{1}{c|}{MobiAct}     
& \multicolumn{1}{c|}{385} & \multicolumn{1}{c|}{1,987} & \multicolumn{1}{c|}{3,991} & \multicolumn{1}{c}{28,028} \\ \midrule
\multicolumn{1}{c|}{MotionSense} 
& \multicolumn{1}{c|}{153}  & \multicolumn{1}{c|}{1,067}  & \multicolumn{1}{c|}{2,148} & \multicolumn{1}{c}{15,092}  \\ \midrule
\multicolumn{1}{c|}{RealWorld}   
& \multicolumn{1}{c|}{520} & \multicolumn{1}{c|}{2,634} & \multicolumn{1}{c|}{5,274} & \multicolumn{1}{c}{36,953} \\ \midrule
\multicolumn{1}{c|}{UCI}         
& \multicolumn{1}{c|}{180}  & \multicolumn{1}{c|}{506}  & \multicolumn{1}{c|}{1,026} & \multicolumn{1}{c}{7,241}  \\ \midrule
\multicolumn{1}{c|}{PAMAP2}       
& \multicolumn{1}{c|}{148}  & \multicolumn{1}{c|}{751}  & \multicolumn{1}{c|}{1,504}  & \multicolumn{1}{c}{10,545}  \\ \bottomrule
\end{tabular}
\label{tab:datasetRatio}
\end{table}
\subsection{Leave-One-Dataset-Out}
\label{subsec:evme}
In order to evaluate the effectiveness of our pre-trained models on unseen data, we introduce a new evaluation strategy called "Leave-One-Dataset-Out" (LODO). The data flow of this approach is depicted in Figure~\ref{fig:LODO}.
This strategy takes into account our multi-dataset setting and provides a better assessment of the model's ability to transfer knowledge to an unseen target domain.
At each fold, one dataset is considered as \textit{left-out dataset}, while the \textit{remaining} ones are used to create a pre-trained model. Specifically, the pre-training phase includes the pre-processing steps depicted in Section~\ref{subsec:combination}  
For the partitioning, these datasets are class-stratified partitioned in a subject-wise manner, with 90\% of the data used for training and 10\% for validation.
The training data from all remaining datasets are then merged (we will refer to this as the source/pre-training dataset), to learn the pre-trained model. Similarly, the validation data from all remaining datasets are used as the validation set during the pre-training process.
The left-out dataset (which we will refer to as the target dataset) is partitioned subject-wise with 20\% set aside for testing and 10\% for validation. The remaining data is used to create various training partitions simulating different labeled data scarcity scenarios for fine-tuning. Six ratios of fine-tuning data are considered: 0\%, 1\%, 5\%, 10\%, and 70\%. The 0\% ratio depicts a scenario where there is no data for fine-tuning, and it is useful to assess the capability of the pre-trained model to transfer knowledge to the target dataset. 
The 1\% ratio assesses the model's performance when there is minimal training data. The 5\% and 10\% ratios demonstrate learning scenarios with limited data availability, while the full 70\% ratio represents a conventional training pipeline.  Table~\ref{tab:datasetRatio} shows the approximate amount of samples regarding each fine-tuning ratio for each of the considered datasets. We observed that, in extreme cases, the PAMAP2 and UCI datasets will have respectively less than 3 and 6 minutes of training data considering the 1\% ratio. 

\subsection{Experimental Setup}
\label{subsec:expset}
Our experiments were conducted on a high-performance computing cluster with Intel Cascade Lake 6248 processors, 192GB of memory, and Nvidia Tesla V100 SXM2 16GB GPUs. All models were developed using TensorFlow~\cite{abadi2016tensorflow}.

For each experiment, we learn the pre-trained model by using $200$ epochs with a batch size of $128$ on the combined datasets mentioned in the previous section. We saved checkpoints of the pre-trained model during training, and then selected the one that achieved the highest accuracy on the validation set as the final pre-trained model. 
In the fine-tuning phase, we adopted $200$ epochs with a batch size of $64$ on the left-out datasets with their representative training scenarios mentioned in the previous section. The evaluation of the test set was performed on the fine-tuned model that achieved the highest accuracy on the validation set.
We adopted Adam as the optimizer with a learning rate of $0.0005$ for all experiments. The hyper-parameters of each specific network follow the standard configuration proposed by the authors of the corresponding papers. We used the macro F1 score metric to measure the performance of the classification task.


\section{Results}
\label{results}

In this section, we present the results of our evaluation using the evaluation methodology proposed in~\ref{subsec:evme}.

\subsection{Main results}

\begin{figure}[h!]
   \centering
       \includegraphics[width=8cm]{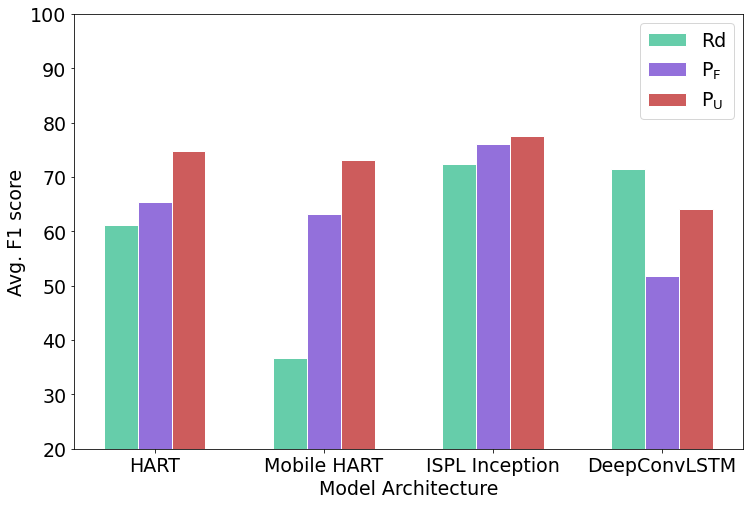}
   \caption{The average F1-score obtained with the considered model architectures by using 1\% of fine-tuning data}
   \label{fig:1_perc_ft}
\end{figure}

We first considered the worst-case labeled data scarcity scenario, where only $1\%$ of fine-tuning data is available in the target domain.
Figure~\ref{fig:1_perc_ft} shows the F1 score obtained on each model with the different fine-tuning strategies, by averaging the recognition rate at each fold.

We observe that pre-training is particularly effective for transformers-based architectures. For example, the F1 score of the randomly initialised HART model $Rd$ is only $61\%$, whereas $P_U$ achieves $74\%$ of F1 score. With MobileHART, the benefits of pre-training are even more noticeable, with a $37\%$ F1 score improvement when using $P_U$ compared to $Rd$. These results suggest that the complexity of transformers-based networks requires a large amount of training data to extract meaningful features, leading to inadequate classification rates with randomly initialised models and limited labeled data in the target domain. Thus, our pre-training method enables transformers to learn high-level and general-purpose features that work well on unseen data, which explains why they reach a high F1 score.

Considering ISPL~Inception, the pre-trained model $P_U$ outperforms $Rd$ on average by $5\%$ of F1 score. In this case, the advantage of pre-training is particularly evident on some datasets (e.g., $\approx +10\%$ when UCI and PAMAP2 are target datasets) and negligible in others (e.g., $\approx +0.4\%$ on MobiAct). This is likely due to the fact that the network underlying ISPL~Inception is less complex than transformers and, in some cases, few training samples from the target domain are sufficient for model convergence. 
On the other hand, we observed that our pre-training strategy has a negative effect on DeepConvLSTM, where the F1 score obtained by fine-tuning $Rd$ outperforms the one generated with $P_U$ and $P_F$. 
This is likely due to the fact that DeepConvLSTM exhibited overfitting behaviors during pre-training, with a negative impact on generalization capabilities on unseen data.

\subsection{More detailed results}

In the following, we provide more details about these results, discussing the pros and cons of the proposed pre-training and fine tuning approach with each of the considered model architectures.

\subsubsection{ISPL Inception}

Table~\ref{tab:ispl_inception} shows the outcomes obtained by using the ISPL~Inception with diverse ratios of fine-tuning data. When fine-tuning is not performed (i.e., fine-tuning ratio of $0\%$), $P_{U}$ achieves an average F1-score of approximately $0.16$ across all target datasets.  This result suggests that, despite pre-training the model with a large number of diverse datasets, this model generally struggles to generalize well on data coming from unseen datasets. 

\begin{figure}[h!]
	\centering

 \centering
	\subfloat[HHAR]{\includegraphics[scale=0.18]{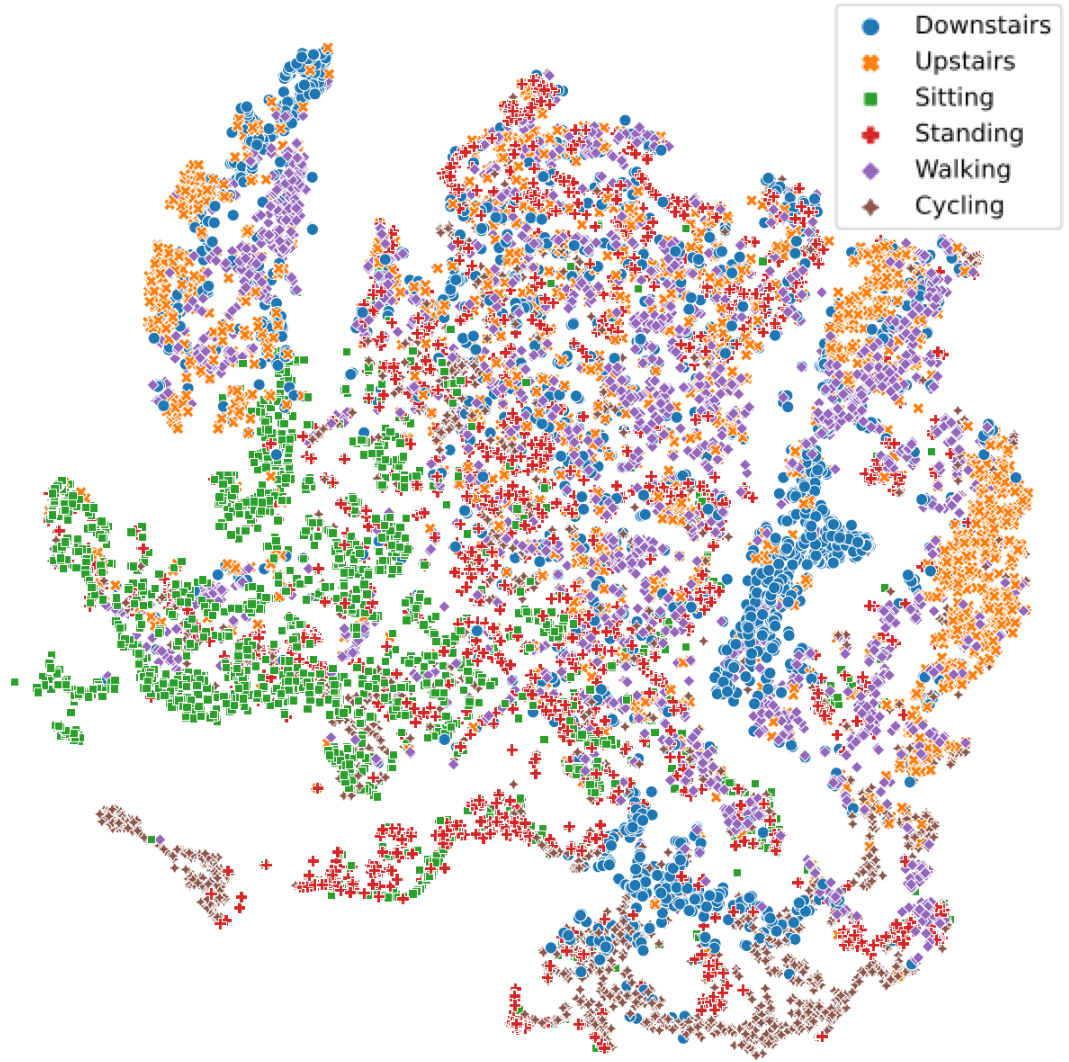}
		\label{fig:embeddings_HHAR}}
	\hfil
	\subfloat[UCI]{	\includegraphics[scale=0.18]{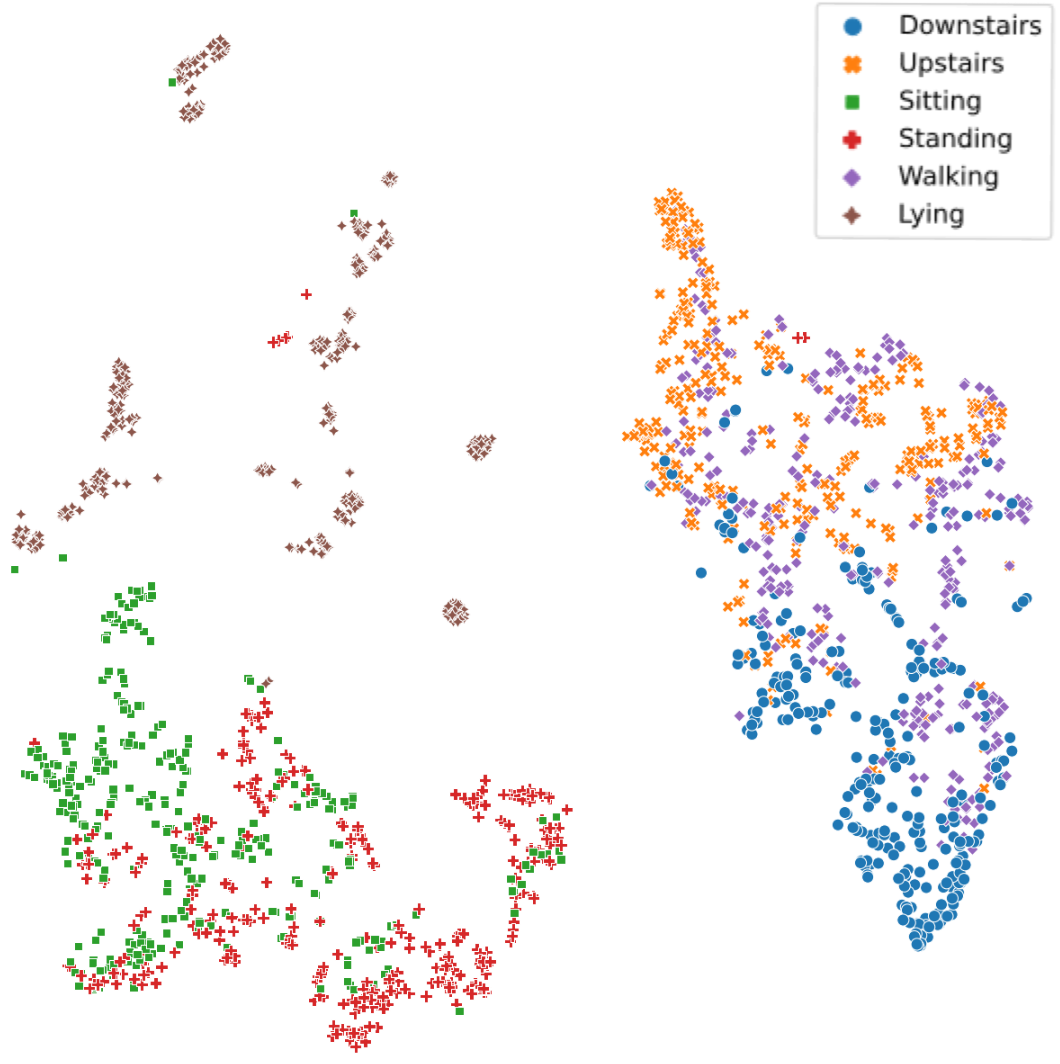}	\label{fig:embeddings_motion_sense}}
	\caption{ T-SNE visualization of the pre-trained ISPL~Inception model's embeddings on HHAR and UCI datasets. The HHAR embeddings are mixed and cluttered, while the ones from UCI show a coarse separation among classes.}
	\label{fig:DeepConv_embeddings}
\end{figure}

Nonetheless, as we previously mentioned, the effectiveness of pre-training varies depending on the specific of the target domain dataset. For example, the F1 score for HHAR and MobiAct with 0\% fine-tuning is below 10\%, while for RealWorld and UCI, it exceeds 20\%. This observation is confirmed by analysing the T-SNE visual representation of the embeddings in Figure~\ref{fig:DeepConv_embeddings}. Here, the embeddings from HHAR are cluttered and poorly separated, while those for UCI exhibit clear separations between static and dynamic activities. 
As previously discussed, the ISPL~Inception model has the ability to quickly learn meaningful features from a small amount of target data, which leads to the recognition rate significantly improving when only 1\% of fine-tuning data is used. Similarly to what we observed with $0\%$, also in this case the advantage of pre-training is significant depending on the specific target dataset. 


As the fine-tuning ratio increases, the benefits of pre-training decrease accordingly, confirming that pre-training is particularly important for labeled data scarcity scenarios.


\begin{table}[]
\footnotesize
\caption{ISPL Inception evaluated using LODO}
\resizebox{\columnwidth}{!}{
\label{tab:ispl_inception}
\begin{tabular}{@{}llcccccccccccc@{}}
\toprule
\multicolumn{1}{c}{\textbf{}} &
   &
  \textbf{} &
  \textbf{} &
  \multicolumn{1}{l}{} &
  \multicolumn{1}{l}{} &
  \multicolumn{1}{l}{} &
  \multicolumn{1}{l}{ft ratio} &
  \multicolumn{1}{l}{} &
  \multicolumn{1}{l}{} &
  \multicolumn{1}{l}{} &
  \multicolumn{1}{l}{} &
  \multicolumn{1}{l}{} &
  \multicolumn{1}{l}{} \\ \midrule
\multicolumn{1}{c|}{\textbf{}} &
  \multicolumn{1}{l|}{\textbf{0\%}} &
  \textbf{} &
  \textbf{1\%} &
  \multicolumn{1}{c|}{} &
   &
  5\% &
  \multicolumn{1}{c|}{} &
   &
  10\% &
  \multicolumn{1}{c|}{} &
   &
  All Train &
  \multicolumn{1}{c}{} \\ \midrule
\multicolumn{1}{l|}{\textbf{\begin{tabular}[c]{@{}l@{}}Target\\ dataset\end{tabular}}} &
  \multicolumn{1}{l|}{${P_{U}}$} &
  Rd &
  ${P_{F}}$ &
  \multicolumn{1}{c|}{${P_{U}}$} &
  Rd &
  ${P_{F}}$ &
  \multicolumn{1}{c|}{${P_{U}}$} &
  Rd &
  ${P_{F}}$ &
  \multicolumn{1}{c|}{${P_{U}}$} &
  Rd &
  ${P_{F}}$ &
  \multicolumn{1}{c}{$P_{U}$} \\ \midrule

\multicolumn{1}{l|}{HHAR} &
\multicolumn{1}{l|}{09.00} &
  80.59  &
  83.33  &
  \multicolumn{1}{c|}{86.12 } &
  91.26 &
  92.04  &
  \multicolumn{1}{c|}{92.39} &
  93.24 &
  93.25 &
  \multicolumn{1}{c|}{94.20} &
  95.70 &
  95.28 &
  \multicolumn{1}{c}{95.34} \\ \midrule
\multicolumn{1}{l|}{MobiAct} &
\multicolumn{1}{l|}{06.67} &
  57.46 &
  59.09 &
  \multicolumn{1}{c|}{57.83 } &
  85.50  &
  83.67&
  \multicolumn{1}{c|}{83.98 } &
  87.68  &
  88.09 &
  \multicolumn{1}{c|}{88.28 } &
  89.63 &
  90.32 &
  \multicolumn{1}{c}{88.53} \\ \midrule

\multicolumn{1}{l|}{MotionSense} &
\multicolumn{1}{l|}{12.67} &
  83.46 &
  83.25 &
  \multicolumn{1}{c|}{84.14} &
  95.34 &
  95.9 &
  \multicolumn{1}{c|}{95.58} &
   96.78 &
  96.73 &
  \multicolumn{1}{c|}{97.36} &
  98.06 &
  98.13 &
  \multicolumn{1}{c}{98.07} \\ \midrule
\multicolumn{1}{l|}{RealWorld} &

\multicolumn{1}{l|}{21.26} &
  81.98 &
  85.25 &
  \multicolumn{1}{c|}{85.90 } &
  87.05 &
  88.17 &
  \multicolumn{1}{c|}{88.35} &
  88.15 &
  89.72  &
  \multicolumn{1}{c|}{88.99 } &
  91.02 &
  91.31 &
  \multicolumn{1}{c}{91.11} \\ \midrule

\multicolumn{1}{l|}{UCI} &

\multicolumn{1}{l|}{27.41} &
  76.06 &
  86.86 &
  \multicolumn{1}{c|}{87.13 } &
  91.14 &
  94.20  &
  \multicolumn{1}{c|}{94.08} &
  92.58 &
  95.81&
  \multicolumn{1}{c|}{95.42} &
  97.45 &
  98.14  &
  \multicolumn{1}{c}{97.76} \\  \midrule
\multicolumn{1}{l|}{PAMAP2} &

\multicolumn{1}{l|}{19.61} &
  54.99 &
  58.76 &
  \multicolumn{1}{c|}{64.07 } &
  72.67 &
  72.55 &
  \multicolumn{1}{c|}{71.73 } &
  76.21&
  74.64 &
  \multicolumn{1}{c|}{74.21} &
  74.09 &
  74.16 &
  \multicolumn{1}{c}{74.17} \\ \midrule

\multicolumn{1}{l}{} &
\multicolumn{1}{l}{} & & &
\multicolumn{1}{c}{}& & &
\multicolumn{1}{c}{}& & &
\multicolumn{1}{c}{} & & &
\multicolumn{1}{c}{} \\ 
  
  \multicolumn{1}{l|}{\textbf{AVG}} &
\multicolumn{1}{l|}{16.10} &
  72.42 &
  76.09 &
  \multicolumn{1}{c|}{77.53} &
  87.16 &
  87.76 &
  \multicolumn{1}{c|}{87.69} &
  89.11 &
 89.71 &
  \multicolumn{1}{c|}{89.74} &
  90.99 &
  91.22 &
  \multicolumn{1}{c}{90.83} \\ \bottomrule
\end{tabular}
}
\end{table}

\subsubsection{Transformer-based models}

Table~\ref{tab:HART} shows the results achieved with HART, while Table \ref{tab:mobile_HART}~illustrates the outcomes generated by MobileHART. Overall, the results are consistent with those obtained with the ISPL Inception network, but some important differences should be noted.

Firstly, the F1 scores achieved with HART and MobileHART by fine-tuning the randomly initialized model $Rd$ with a small percentage of data (i.e., 1\% and 5\%) are generally lower than those obtained with the ISPL Inception network. This can be attributed to the higher complexity of the transformer architecture, which requires more training samples to extract representative features from the target dataset.

On the other hand, the performances in terms of F1 score obtained by fine-tuning $P_{U}$ and $P_{F}$ in these scenarios are significantly better than the F1 score achieved with $Rd$. 

Furthermore, the advantages of pre-training are noticeable even when using $5\%$ and $10\%$ of the fine-tuning data. 




\begin{table}[]
\footnotesize
\caption{HART evaluated using LODO}
\resizebox{\columnwidth}{!}{
\begin{tabular}{@{}llcccccccccccc@{}}
\toprule
\multicolumn{1}{c}{\textbf{}} &
   &
  \textbf{} &
  \textbf{} &
  \multicolumn{1}{l}{} &
  \multicolumn{1}{l}{} &
  \multicolumn{1}{l}{} &
  \multicolumn{1}{l}{ft ratio} &
  \multicolumn{1}{l}{} &
  \multicolumn{1}{l}{} &
  \multicolumn{1}{l}{} &
  \multicolumn{1}{l}{} &
  \multicolumn{1}{l}{} &
  \multicolumn{1}{l}{} \\ \midrule
\multicolumn{1}{c|}{\textbf{}} &
  \multicolumn{1}{c|}{\textbf{0\%}} &
  \textbf{} &
  \textbf{1\%} &
  \multicolumn{1}{c|}{} &
   &
  5\% &
  \multicolumn{1}{c|}{} &
   &
  10\% &
  \multicolumn{1}{c|}{} &
   &
  All Train &
  \multicolumn{1}{c}{} \\ \midrule
\multicolumn{1}{l|}{\textbf{\begin{tabular}[c]{@{}l@{}}Target\\ dataset\end{tabular}}} &
  \multicolumn{1}{l|}{${P_{U}}$} &
  Rd &
  $P_{F}$ &
  \multicolumn{1}{c|}{$P_{U}$} &
  Rd &
  $P_{F}$ &
  \multicolumn{1}{c|}{$P_{U}$} &
  Rd &
  $P_{F}$ &
  \multicolumn{1}{c|}{$P_{U}$} &
  Rd &
  $P_{F}$ &
  \multicolumn{1}{c}{$P_{U}$} \\ \midrule

\multicolumn{1}{l|}{HHAR} &
\multicolumn{1}{l|}{08.11} &
  65.28 &
  64.72 &
  \multicolumn{1}{c|}{66.55} &
  83.48 &
  83.69 &
  \multicolumn{1}{c|}{84.66} &
  89.82 &
  88.16 &
  \multicolumn{1}{c|}{89.42} &
  93.30 &
  92.75 &
  \multicolumn{1}{c}{93.86} \\ \midrule
\multicolumn{1}{l|}{MobiAct} &
\multicolumn{1}{l|}{07.81} &
  43.70 &
  50.56 &
  \multicolumn{1}{c|}{67.99 } &
  77.69 &
  80.40 &
  \multicolumn{1}{c|}{80.05} &
  83.16 &
  82.86  &
  \multicolumn{1}{c|}{83.78} &
  87.67 &
  87.57 &
  \multicolumn{1}{c}{87.33} \\ \midrule
\multicolumn{1}{l|}{MotionSense} &

\multicolumn{1}{l|}{14.04} &
  70.91 &
  74.23 &
  \multicolumn{1}{c|}{87.93} &
  91.66 &
  92.80 &
  \multicolumn{1}{c|}{93.17 } &
  93.56 &
  94.88  &
  \multicolumn{1}{c|}{95.49} &
  97.45 &
  97.92 &
  \multicolumn{1}{c}{97.84} \\ \midrule

\multicolumn{1}{l|}{RealWorld} &
\multicolumn{1}{l|}{32.68} &
  77.08 &
  75.38 &
  \multicolumn{1}{c|}{85.75} &
  85.83 &
  87.74 &
  \multicolumn{1}{c|}{87.89} &
  87.45 &
  88.99 &
  \multicolumn{1}{c|}{89.33 } &
  91.30 &
  91.81 &
  \multicolumn{1}{c}{91.64} \\ \midrule

\multicolumn{1}{l|}{UCI} &
\multicolumn{1}{l|}{26.69} &
  61.00 &
  70.22 &
  \multicolumn{1}{c|}{84.46} &
  88.83 &
  79.82 &
  \multicolumn{1}{c|}{93.27 } &
    90.20 &
  95.32 &
  \multicolumn{1}{c|}{ 95.06} &
  96.40 &
  96.70 &
  \multicolumn{1}{c}{96.74} \\ \midrule
\multicolumn{1}{l|}{PAMAP2} &
\multicolumn{1}{l|}{15.39} &
  49.17 &
  57.82 &
  \multicolumn{1}{c|}{56.31} &
  63.04 &
  64.82 &
  \multicolumn{1}{c|}{64.62} &
  64.42 &
  66.82 &
  \multicolumn{1}{c|}{66.16} &
  69.10 &
  68.13 &
  \multicolumn{1}{c}{67.91} \\ \midrule

\multicolumn{1}{l}{} &
\multicolumn{1}{l}{} & & &
\multicolumn{1}{c}{}& & &
\multicolumn{1}{c}{}& & &
\multicolumn{1}{c}{} & & &
\multicolumn{1}{c}{} \\ 
  
  \multicolumn{1}{l|}{\textbf{AVG}} &
\multicolumn{1}{l|}{17.45} &
  61.19 &
  65.49 &
  \multicolumn{1}{c|}{74.83} &
  81.76 &
  81.55 &
  \multicolumn{1}{c|}{83.94} &
  84.77 &
  86.17 &
  \multicolumn{1}{c|}{86.54} &
  89.20 &
  89.15 &
  \multicolumn{1}{c}{89.22} \\ \bottomrule
\end{tabular}
}
\label{tab:HART}
\end{table}

\begin{table}[]
\centering
\caption{MobileHART evaluated using LODO}
\resizebox{\columnwidth}{!}{
\footnotesize
\begin{tabular}{@{}llcccccccccccc@{}}
\toprule
\multicolumn{1}{c}{\textbf{}} &
   &
  \textbf{} &
  \textbf{} &
  \multicolumn{1}{l}{} &
  \multicolumn{1}{l}{} &
  \multicolumn{1}{l}{} &
  \multicolumn{1}{l}{ft ratio} &
  \multicolumn{1}{l}{} &
  \multicolumn{1}{l}{} &
  \multicolumn{1}{l}{} &
  \multicolumn{1}{l}{} &
  \multicolumn{1}{l}{} &
  \multicolumn{1}{l}{} \\ \midrule
\multicolumn{1}{c|}{\textbf{}} &
  \multicolumn{1}{l|}{\textbf{0\%}} &
  \textbf{} &
  \textbf{1\%} &
  \multicolumn{1}{c|}{} &
   &
  5\% &
  \multicolumn{1}{c|}{} &
   &
  10\% &
  \multicolumn{1}{c|}{} &
   &
  All Train &
  \multicolumn{1}{c}{} \\ \midrule
\multicolumn{1}{l|}{\textbf{\begin{tabular}[c]{@{}l@{}}Target\\ dataset\end{tabular}}} &
  \multicolumn{1}{l|}{${P_{U}}$} &
  Rd &
  $P_{F}$ &
  \multicolumn{1}{c|}{$P_{U}$} &
  Rd &
  $P_{F}$ &
  \multicolumn{1}{c|}{$P_{U}$} &
  Rd &
  $P_{F}$ &
  \multicolumn{1}{c|}{$P_{U}$} &
  Rd &
  $P_{F}$ &
  \multicolumn{1}{c}{$P_{U}$} \\ \midrule
\multicolumn{1}{l|}{HHAR} &

\multicolumn{1}{l|}{06.27} &
  78.02 &
  60.18 &
  \multicolumn{1}{c|}{71.03 } &
  92.76  &
  91.26  &
  \multicolumn{1}{c|}{92.05 } &
  94.9  &
  93.99 &
  \multicolumn{1}{c|}{94.06 } &
  97.31 &
  97.07 &
  \multicolumn{1}{c}{96.63} \\ \midrule
\multicolumn{1}{l|}{MobiAct} &
\multicolumn{1}{l|}{06.84} &
  47.05 &
  44.42 &
  \multicolumn{1}{c|}{53.50 } &
  82.47  &
  75.38  &
  \multicolumn{1}{c|}{80.93} &
  86.53  &
  84.29 &
  \multicolumn{1}{c|}{85.29 } &
  89.19 &
  89.22 &
  \multicolumn{1}{c}{90.18} \\ \midrule
\multicolumn{1}{l|}{MotionSense} &

\multicolumn{1}{l|}{16.26} &
  10.92  &
  68.34 &
  \multicolumn{1}{c|}{85.92} &
  92.39 &
  92.18  &
  \multicolumn{1}{c|}{93.06 } &
  95.85  &
  96.44  &
  \multicolumn{1}{c|}{96.89 } &
  98.26 &
  97.82 &
  \multicolumn{1}{c}{98.16} \\ \midrule

\multicolumn{1}{l|}{RealWorld} &
\multicolumn{1}{l|}{33.52} &
  66.97 &
  83.61   &
  \multicolumn{1}{c|}{84.99  } &
  85.23 &
  87.42 &
  \multicolumn{1}{c|}{87.60} &
  88.64  &
  89.21   &
  \multicolumn{1}{c|}{89.18 } &
  91.30 &
  91.26 &
  \multicolumn{1}{c}{91.13} \\ \midrule
\multicolumn{1}{l|}{UCI} &
\multicolumn{1}{l|}{20.65} &
  13.44  &
  67.25 &
  \multicolumn{1}{c|}{84.65} &
  88.64&
  92.46 &
  \multicolumn{1}{c|}{94.18} &
  90.76   &
  92.95  &
  \multicolumn{1}{c|}{95.17} &
  97.44 &
  97.40 &
  \multicolumn{1}{c}{98.24} \\ \midrule
\multicolumn{1}{l|}{PAMAP2} &

\multicolumn{1}{l|}{12.58} &
  04.25 &
  55.7  &
  \multicolumn{1}{c|}{ 58.53 } &
  60.03 &
  65.73 &
  \multicolumn{1}{c|}{68.48 } &
  65.01  &
  67.70  &
  \multicolumn{1}{c|}{70.34} &
  71.10 &
  71.89&
  \multicolumn{1}{c}{73.98} \\ \midrule

\multicolumn{1}{l}{} &
\multicolumn{1}{l}{} & & &
\multicolumn{1}{c}{}& & &
\multicolumn{1}{c}{}& & &
\multicolumn{1}{c}{} & & &
\multicolumn{1}{c}{} \\ 
  
  \multicolumn{1}{l|}{\textbf{AVG}} &
\multicolumn{1}{l|}{16.02} &
  36.78 &
  63.25 &
  \multicolumn{1}{c|}{73.10} &
  83.59&
  84.07 &
  \multicolumn{1}{c|}{86.05} &
  86.95 &
  87.43 &
  \multicolumn{1}{c|}{88.49} &
  90.77 &
  90.78 &
  \multicolumn{1}{c}{91.39} \\ \bottomrule
\end{tabular}
}
\label{tab:mobile_HART}
\end{table}

\subsubsection{DeepConvLSTM}

Table \ref{tab:deep_conv_LSTM} shows that, in general, DeepConvLSTM does not benefit from pre-training contrary to the other model architectures considered in this study. Specifically, the randomly initialized model $Rd$ outperformed the pre-trained models $P_U$ and $P_F$ when a limited percentage of fine-tuning data was used (i.e., 1\% and 5\%).
This behavior may be due to the fact that the DeepConvLSTM has significantly fewer parameters compared to the other networks (see Table \ref{tab:num_parameters} for details), and therefore struggles to extract high-level features that generalize well over heterogeneous datasets.
\begin{table}[h!]
\footnotesize
\centering
\caption{Models with their parameters and FLOPS}
\begin{tabular}{|l|r|r|}
\hline
\textbf{Model}              & \multicolumn{1}{l|}{\textbf{Parameters}} & \multicolumn{1}{l|}{\textbf{FLOPS}} \\ \hline
ISPL Inception \cite{ronald2021isplinception} & 1,327,714    & 338,528,096   \\ \hline
DeepConvLSTM \cite{ordonez2016deep} & 457,546   & 86,184,508   \\ \hline
HART \cite{ek2022lightweight} & 1,445,918  & 15,212,636   \\ \hline
 MobileHART \cite{ek2022lightweight} & 2,542,942   & 19,809,292    \\ \hline
\end{tabular}\label{tab:num_parameters}
\end{table}
Nonetheless, there is only one case where our pre-training strategy is effective for DeepConvLSTM, and that is when UCI is used as a target dataset.
For these reasons, DeepConvLSTM resulted in not a particularly suitable model for the considered transfer-learning task. 

\begin{table}[]
\centering
\caption{DeepConvLSTM evaluated using LODO}
\resizebox{\columnwidth}{!}{

\begin{tabular}{@{}llcccccccccccc@{}}
\toprule
\multicolumn{1}{c}{\textbf{}} &
   &
  \textbf{} &
  \textbf{} &
  \multicolumn{1}{l}{} &
  \multicolumn{1}{l}{} &
  \multicolumn{1}{l}{} &
  \multicolumn{1}{l}{ft ratio} &
  \multicolumn{1}{l}{} &
  \multicolumn{1}{l}{} &
  \multicolumn{1}{l}{} &
  \multicolumn{1}{l}{} &
  \multicolumn{1}{l}{} &
  \multicolumn{1}{l}{} \\ \midrule
\multicolumn{1}{c|}{\textbf{}} &
  \multicolumn{1}{l|}{\textbf{0\%}} &
  \textbf{} &
  \textbf{1\%} &
  \multicolumn{1}{c|}{} &
   &
  5\% &
  \multicolumn{1}{c|}{} &
   &
  10\% &
  \multicolumn{1}{c|}{} &
   &
  All Train &
  \multicolumn{1}{c}{} \\ \midrule
\multicolumn{1}{l|}{\textbf{\begin{tabular}[c]{@{}l@{}}Target\\ dataset\end{tabular}}} &
  \multicolumn{1}{l|}{${P_{U}}$} &
  Rd &
  $P_{F}$ &
  \multicolumn{1}{c|}{$P_{U}$} &
  Rd &
  $P_{F}$ &
  \multicolumn{1}{c|}{$P_{U}$} &
  Rd &
  $P_{F}$ &
  \multicolumn{1}{c|}{$P_{U}$} &
  Rd &
  $P_{F}$ &
  \multicolumn{1}{c}{$P_{U}$} \\ \midrule


\multicolumn{1}{l|}{HHAR} &
\multicolumn{1}{l|}{05.58} &
  80.37 &
  60.00 &
  \multicolumn{1}{c|}{66.26} &
  83.28 &
  79.03 &
  \multicolumn{1}{c|}{82.12} &
  86.15 &
  87.71 &
  \multicolumn{1}{c|}{89.42} &
  88.48 &
  94.04 &
  \multicolumn{1}{c}{94.87} \\ \midrule


\multicolumn{1}{l|}{MobiAct} &
\multicolumn{1}{l|}{06.55} &
  51.94 &
 28.21 &
  \multicolumn{1}{c|}{42.14} &
  85.46 &
  72.38 &
  \multicolumn{1}{c|}{77.71} &
  85.46 &
   79.79 &
  \multicolumn{1}{c|}{81.11} &
  88.72 &
  87.47 &
  \multicolumn{1}{c}{87.62} \\ \midrule

\multicolumn{1}{l|}{MotionSense} &
\multicolumn{1}{l|}{14.71} &
  87.67  &
  42.67&
  \multicolumn{1}{c|}{67.54} &
  95.60 &
  76.6 &
  \multicolumn{1}{c|}{91.64} &
  96.71  &
  94.60&
  \multicolumn{1}{c|}{95.26} &
    98.17 &
    98.01 &
  \multicolumn{1}{c}{98.20} \\ \midrule


\multicolumn{1}{l|}{RealWorld} &
\multicolumn{1}{l|}{28.07} &
  84.64  &
    66.93  &
  \multicolumn{1}{c|}{ 77.39} &
    88.13  &
    86.22 &
  \multicolumn{1}{c|}{ 87.62} &
    89.50  &
    89.48 &
  \multicolumn{1}{c|}{ 89.39} &
    91.28 &
    91.28 &
  \multicolumn{1}{c}{ 91.52} \\ \midrule


\multicolumn{1}{l|}{UCI} &
\multicolumn{1}{l|}{37.13} &
  68.88 &
  64.65  &
  \multicolumn{1}{c|}{79.43} &
    89.81 &
    76.81 &
  \multicolumn{1}{c|}{91.05} &
    92.29  &
    93.3  &
  \multicolumn{1}{c|}{95.52} &
  97.53 &
  96.50 &
  \multicolumn{1}{c}{ 97.20} \\ \midrule



\multicolumn{1}{l|}{PAMAP2} &
\multicolumn{1}{l|}{20.16} &
  55.47 &
  47.81 &
  \multicolumn{1}{c|}{52.35} &
  68.69 &
  63.54 &
  \multicolumn{1}{c|}{64.56} &
  66.36 &
   66.93 &
  \multicolumn{1}{c|}{67.21} &
  71.84 &
   68.95 &
  \multicolumn{1}{c}{68.91} \\ \midrule

\multicolumn{1}{l}{} &
\multicolumn{1}{l}{} & & &
\multicolumn{1}{c}{}& & &
\multicolumn{1}{c}{}& & &
\multicolumn{1}{c}{} & & &
\multicolumn{1}{c}{} \\ 
  
  \multicolumn{1}{l|}{\textbf{AVG}} &
\multicolumn{1}{l|}{18.70} &
  71.50 &
 51.71 &
  \multicolumn{1}{c|}{64.19} &
  85.16 &
  75.76 &
  \multicolumn{1}{c|}{82.45} &
  86.08 &
  85.30 &
  \multicolumn{1}{c|}{86.32} &
  89.34 &
  89.38 &
  \multicolumn{1}{c}{89.72} \\ \bottomrule
\end{tabular}
}
\label{tab:deep_conv_LSTM}
\end{table}


\section{Discussion}
\label{discussion}


The beneficial impact of the pre-training varies across different architectures and target datasets. 
This show that it is important to evaluate pre-training on different datasets and different types of models. 
Transformer-based architectures such as HART and MobileHART benefit more from the pre-training as opposed to smaller networks such as the DeepConvLSTM, which obtained ranging benefits from little to worsening performances when compared to starting randomly instead. As transformer-based networks generally require more training and data to start converging to superior performances, we argue that model initializing on other datasets is essential for this type of architecture. Contrarily, smaller networks are more prone to overfitting when starting with an initialized model that has been trained on a large amount of data.

In terms of datasets, reported performances vary. Fine-tuning the pre-trained model using datasets such as  MotionSense, RealWorld, UCI, and PAMAP2 generally showed improved recognition rates. However, when the target dataset is too different from the other combined datasets model, the advantage of pre-training may be limited. For instance, considering the HHAR and MobiAct datasets, our results show that it is particularly challenging to build a pre-trained model capable of generalizing on them. These results are consistent considering all the neural networks by observing the results of transfer without any fine-tuning. 
This is likely due to the numerous devices used and the abundant number of users of the HHAR and MobiAct datasets respectively. 

Finally, the results obtained by fine-tuning the three best model using a frozen feature extractor ($P_F$) are similar to the ones obtained with an entirely fine-tuned pre-trained model ($P_U$). This suggests that the pre-trained feature extractor has learned representations that are sufficiently robust to reach good performances simply by optimizing a classification head. 
Such results are particularly relevant considering training efficiency since the frozen-feature extractor scenario drastically lowers training costs in the fine-tuning stage.






\section{Conclusion}
\label{conclusion}

This paper introduces a novel strategy to mitigate the labeled data scarcity problem for sensor-based HAR, that relies on combining public datasets.
By proposing a new evaluation methodology, 
our results show that pre-training on multiple datasets greatly improves performance in labeled data scarcity scenarios. 
In future work, we aim at improving the pre-training mechanism by learning more robust features. Specifically, we intend to investigate self-supervised learning methods. Indeed, several studies in other domains (e.g. Computer Vision, NLP) have already shown that the trained model is able to learn task-agnostic features~\cite{9462394}.

\section{Acknowledgement}
\label{acknowledgement}
\footnotesize

This work has been partially funded by Naval Group, by MIAI@Grenoble Alpes (ANR-19-P3IA-0003), and granted access to the HPC resources of IDRIS under the allocation 2023-AD011013233R1 made by GENCI.

Part of this research was also supported by projects SERICS (PE00000014) and by project MUSA – Multilayered Urban Sustainability Action,  funded by the European Union – NextGenerationEU, under the National Recovery and Resilience Plan (NRRP) Mission 4 Component 2 Investment Line 1.5: Strengthening of research structures and creation of R\&D “innovation ecosystems”, set up of “territorial leaders in R\&D”.


\bibliographystyle{ieeetr}
\bibliography{references}

\end{document}